\pdfoutput=1

\documentclass[11pt]{article}

\usepackage[]{acl}

\usepackage{times}
\usepackage{latexsym}

\usepackage[T1]{fontenc}

\usepackage[utf8]{inputenc}
\usepackage[most]{tcolorbox}
\usepackage[inkscapelatex=false]{svg}
\usepackage{microtype}
\usepackage{multirow}
\usepackage{graphicx}
\usepackage{booktabs}
\usepackage[skip=5pt]{caption}
\usepackage{inconsolata}
\usepackage{amsmath}
\usepackage{booktabs}
\usepackage{pdfpages}
\newcommand{\ours}{\texttt{ASC}}
\newcommand{\asc}{\texttt{ASC}}
\newcommand{\usc}{\texttt{USC}}
\newcommand{\uscl}{\texttt{USC-LLM}}
\newcommand{\fc}{\texttt{ASC-F}}
\newcommand{\direct}{\texttt{Direct}}
\newcommand{\acf}{\texttt{ACF}}
\newcommand{\fcf}{\texttt{FCF}}
\newcommand{\rc}{\texttt{Random Clusters}}
\newcommand{\rs}{\texttt{Random Sentences}}

\title{Atomic Self-Consistency for Better Long Form Generations}

\author{\hspace{10em}Raghuveer Thirukovalluru,\\\hspace{25em}Duke University\\\hspace{14em}\texttt{\{raghuveer.thirukovalluru, yukun.huang\}@duke.edu},\And
  \hspace{4em}Yukun Huang,\\\And
  \hspace{-6.4em}Bhuwan Dhingra\\\\\hspace{-1em}\texttt{bdhingra@cs.duke.edu}}

\begin{document}
\maketitle
\begin{abstract}

Recent work has aimed to improve LLM generations by filtering out hallucinations, thereby improving the precision of the information in responses. Correctness of a long-form response, however, also depends on the recall of multiple pieces of information relevant to the question. In this paper, we introduce Atomic Self-Consistency (\asc), a technique for improving the recall of relevant information in an LLM response. \asc\ follows recent work, Universal Self-Consistency (\usc) in using multiple stochastic samples from an LLM to improve the long-form response. Unlike \usc\ which only focuses on selecting the best single generation, \asc\ picks authentic subparts from the samples and merges them into a superior composite answer. Through extensive experiments and ablations, we show that merging relevant subparts of multiple samples performs significantly better than picking a single sample.  \asc\ demonstrates significant gains over \usc\ on multiple factoids and open-ended QA datasets - ASQA, QAMPARI, QUEST, ELI5 with ChatGPT and Llama2. Our analysis also reveals untapped potential for enhancing long-form generations using approach of merging multiple samples.
\end{abstract}

\section{Introduction}
Large language models (LLM) with their ability to perform mathematical reasoning \cite{wei2022chain}, planning \cite{ahn2022can}, and generating human-like text \cite{bubeck2023sparks} have become an integral component of many AI systems. Long-form question answering (LFQA) is an important benchmark task whose performance reflects the reliability of these AI systems 
at providing comprehensive and accurate responses to user queries.

\begin{figure}     
     \centering
     \includegraphics[width=0.9\columnwidth]{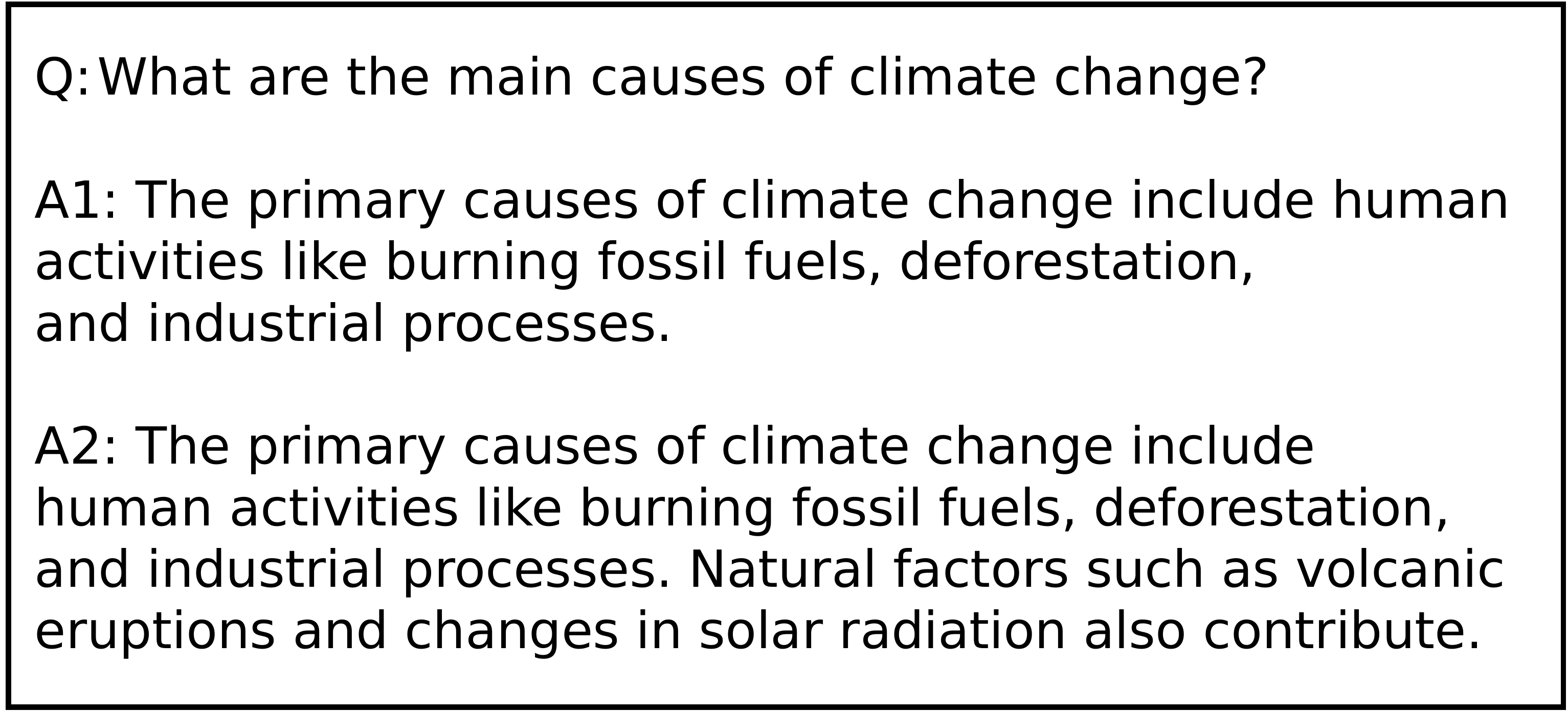}
    \caption{A$_1$: A precise answer. A$_2$: An answer with higher recall of atomic facts relevant to the question Q.  
    }
    \label{fig1}
    \vspace{-1.5em}
\end{figure}

In LFQA, each response comprises multiple pieces of information, described in the literature as atomic facts \cite{min2023factscore}, that collectively contribute to the overall correctness of the answer. Despite various improvements, LLMs are still very prone to producing hallucinatory content such as incorrect atomic facts, especially when the responses are longer \cite{ren2023self}. Recent works on mitigating hallucinations have primarily involved the removal of inaccurate atomic facts from the generated content. While these methods produce responses with high precision over atomic facts, the correctness of the response also depends on the inclusion of all information relevant to the question, i.e., the recall of atomic facts relevant to the question. E.g. In Fig. \ref{fig1}, A$_1$ is a precise response. A$_2$ is a more complete high recall response to Q.

On the other hand, in QA with closed-form answers (such as a math problem with a numeric answer), remarkable improvements in response quality have been achieved by stochastically sampling multiple model responses and then using consistency criteria to select one as the final answer \cite{wang2022self}. Recently, similar efforts were extended to long-form generation. Universal Self Consistency (\usc) \cite{chen2023universal}, is one such example which uses LLMs to determine consistency between model responses. The output of \usc\ is the single most consistent generation among multiple candidate samples from the model.

\begin{figure}     
     \centering
     \includegraphics[width=\columnwidth]{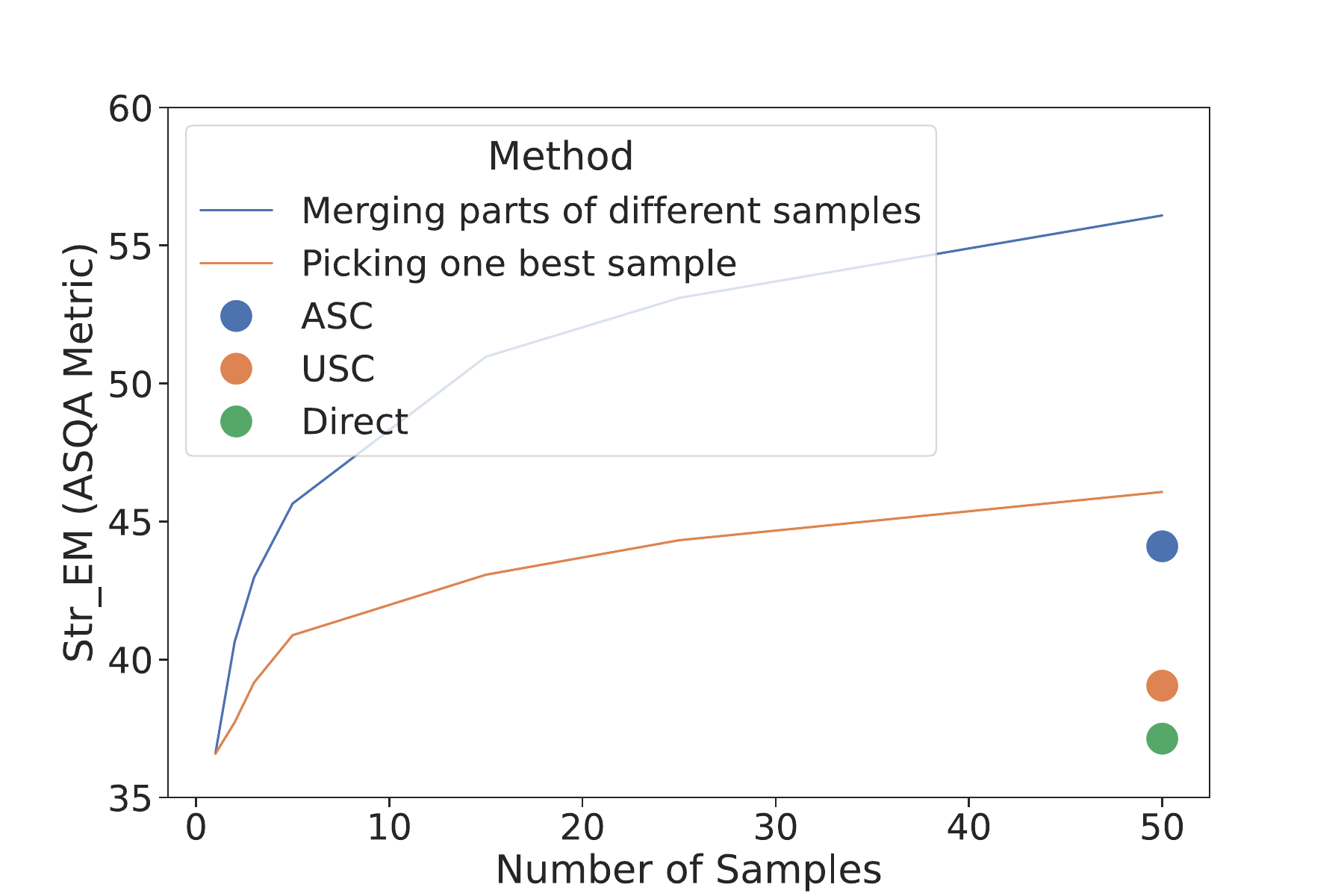}
    \caption{Best possible recall (oracle performance) with increasing number of samples on ASQA(ChatGPT). Merging subparts from multiple samples has a much higher ceiling. \asc\ beats \usc, \direct; almost matches the ceiling performance of picking one best sample.
    }
    \label{fig_pickvsmerge}
    \vspace{-1.7em}
\end{figure}

However, picking a single final answer among the candidate generations might miss out on relevant atomic facts from other candidates and not optimize the recall of information. Further, it is still prone to some atomic hallucinations within the final selected candidate. To overcome these challenges, we propose a simple approach called Atomic Self-Consistency (ASC), which combines authentic atomic facts from multiple candidate responses to generate a superior composite response. To motivate the readers on the potential benefits of this approach, Fig. \ref{fig_pickvsmerge} shows the oracle performance (best possible performance) of picking one single generation vs merging subparts of multiple generations on the ASQA dataset (details in \S \ref{ss:o_recall}). Merging answers from multiple samples have significant performance potential over picking a single answer. \usc\ would not be able to capture this potential. Fig. \ref{fig_pickvsmerge} also shows the performance of \asc\ and other baselines. \asc\ matches the ceiling performance possible by picking the best sample.

\begin{figure*}     
     \centering
     \includegraphics[width=\textwidth]{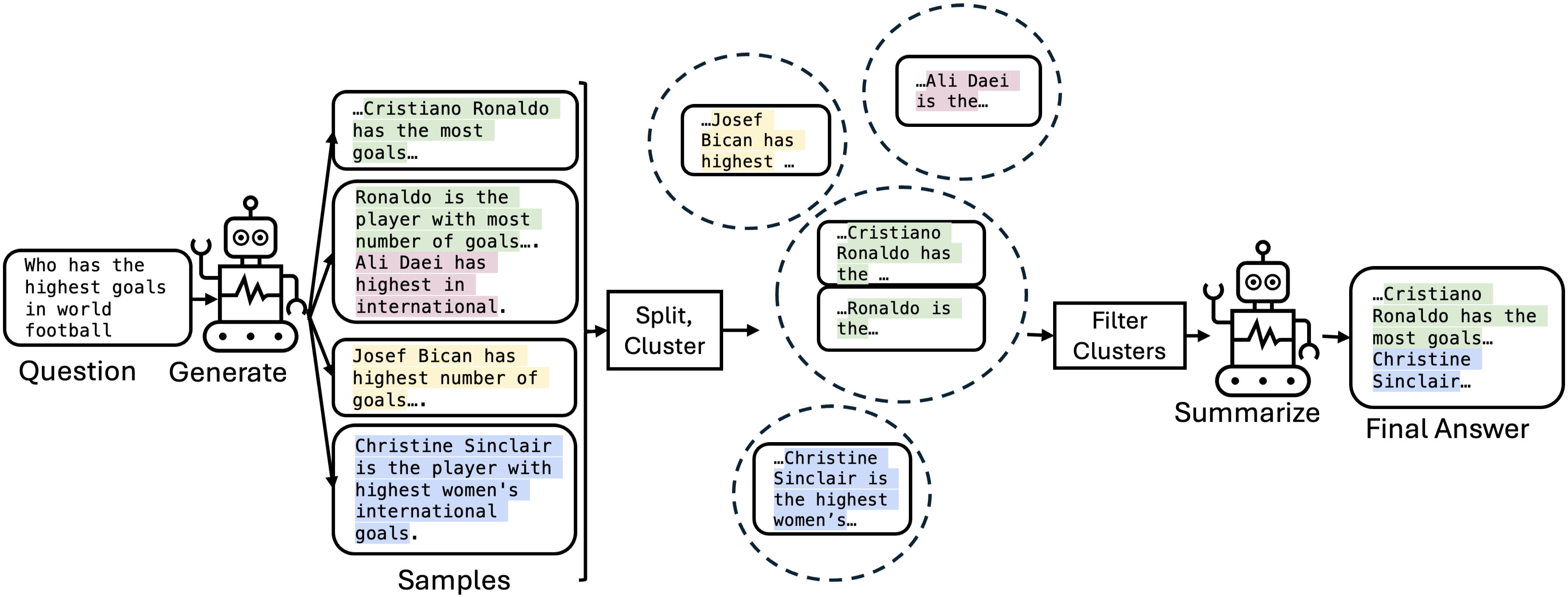}
    \caption{Overall Pipeline proposed. Generated samples are split into smaller parts and clustered. Clusters are then filtered by a consistency based criterion (higher strength clusters are selected while lower strength clusters are removed). Selected cluster representatives are then summarized by an LLM to generate a final answer.}
    \label{fig_method}
    \vspace{-1em}
\end{figure*}

\noindent\textbf{Key Contributions}
1. We introduce a simple and efficient method, \asc, that: (i) clusters atomic parts of multiple candidate generations, (ii) uses a consistency-based criterion to pick the best clusters relevant to the question, and (iii) finally combines them all into a single superior answer. \asc\ operates in black box mode and does not require access to LLM weights or logits.
2. We systematically establish the superiority of combining multiple generations over picking a single generation.
3. We extensively evaluate and show significant performance improvements of \asc\ over \usc\ and other strong baselines on four diverse long-form QA tasks---ASQA, QAMPARI, QUEST and ELI5. We justify the benefits of \asc\ through rigorous ablations.
4. We show strong empirical evidence for minimizing the number of stochastic samples required by \asc. 5. Finally, our analysis reveals untapped potential for enhancing LLMs by merging multiple generations into a superior composite output. This insight underscores a promising avenue for advancing LLM performance further.

\section{Related Work}\label{sec:rel_work}
This related work section first elucidates key methodologies that were used to identify hallucinations and alleviate hallucinations. We further talk about the importance of \textit{consistency} as a measure for correctness and finally talk about how this is used to improve LLM response.

\noindent\textbf{Detecting and alleviating hallucination}: FactScore \cite{min2023factscore} proposed a mechanism to identify atomic facts using an LLM and then individually check each fact's correctness using a retrieval-based solution. CoVe \cite{dhuliawala2023chainofverification} used the LLM to generate multiple verification questions over a candidate generation and prompted an LLM to answer them. Answers to these verifying questions were used to draft a high-precision response. \citet{agrawal2023language} used indirect prompts to verify individual units in list-style answers.

\noindent\textbf{Consistency as a measure for correctness}: Consistency between model responses resulted in performance leaps in mathword problems, code generation, etc \cite{xiong2022self, shi2022natural}. SelfCheckGPT \cite{manakul2023selfcheckgpt} is another work very relevant to \asc\ which detects hallucination in model responses. It verifies the correctness of individual sentences in a generation by measuring their agreement with multiple other stochastic samples by the LLM. It showed the benefit of consistency-based measures to identify sentence-level hallucination, within long-form generations. HaLo \cite{elaraby2023halo} also used consistency-based measures to identify sentence-level hallucinations in a generated text. It further explored techniques like knowledge injection and distillation for alleviating hallucinations. 

\noindent\textbf{Using stochastic samples to improve LLM response}: \usc\ \cite{chen2023universal} uses the consistency-based measure to pick the most consistent individual response among stochastically generated sample responses. It feeds in all responses to the LLM and asks to output the most consistent response. Similar to \usc, \citet{ren2023self} pick one best answer from a list of candidates by a hybrid mechanism which contains a score from the samples posed as a multiple choice question and a score based on self evaluation from the same LLM. 

Despite remarkable improvements, these methods are confined to either filtering atomic facts from a response or picking a single final answer from multiple samples. Our research on the other hand, focuses on combining subparts of multiple stochastic samples to produce higher quality generations.

\section{Methodology}
Given a question $q$, our task is to use an LLM $\mathcal{L}$ to produce an answer
which answers the questions both accurately (i.e., with high precision) and comprehensively
(i.e., with high recall).
We describe the specific metrics used to measure this for each dataset in \S\ref{sec:tasks}.
Let $a_1, a_2, .., a_m$ be $m$ independent samples directly generated by $\mathcal{L}$ when query $q$ is fed to it in a prompt $P$. This work merges consistent subparts from the multiple samples to produce a final answer $a_{\asc}$ in a four-step process. 1. Split: Split each generation into its constituent atomic facts. 2. Cluster: Grouping atomic facts for efficiency 3. Filter: Selecting the most consistent clusters. 4. Summarize: Combine the selected cluster representatives to generate a final answer.  


\subsection{Splitting Generations for Atomic Facts}
As the veracity of atomic facts can be verified, the first step in our approach is to break down candidate generations into atomic facts. A candidate generation to a question might comprise multiple sentences and multiple atomic facts within each sentence.
\citet{min2023factscore} used an InstructGPT model to break down a long-form generation into its atomic facts. While we believe that use of such neural models can produce better quality atomic facts, it is extremely expensive in our setting as this needs to be performed for $m$ different generations per question. Hence, following \citet{arslan2020benchmark, manakul2023selfcheckgpt, liu2023evaluating}, we used individual sentences within a candidate generation as its atomic facts. We used sentence tokenization models \cite{bird2009natural} to split the generation into its constituent sentences (atomic facts).

\subsection{Clustering Atomic Facts}
Each of the $m$ generations typically contains multiple individual sentences/atomic facts, say $k$ on average. Each of these units needs to be verified for relevance and truthfulness before inclusion in the final answer. However, evaluating each atomic unit either by external sources like retrieval or by prompting $\mathcal{L}$ would require $m*k$ steps and prove very expensive. Note that multiple of these atomic units share content because they were generated addressing the same question. We hence perform a round of clustering to collect atomic facts conveying the same meaning. Despite being cubic in computational complexity, we found agglomerative clustering \cite{pedregosa2011scikit} of sentence embeddings obtained from SimCSE \cite{gao2021simcse} has less overhead and is much faster compared to using retrieval/LLM calls to filter them.

Given the clusters, verifying only the representatives of each cluster would contribute to substantial savings in compute and time. We chose the longest sentence in each cluster as its representative.

\subsection{Filtering Clusters}\label{c_c}
The objective now is to identify and eliminate inaccurate atomic facts, ensuring that only valid and reliable atomic facts are retained for the final answer generation in the subsequent steps. Multiple methods have been used in the literature to verify facts, e.g., retrieval-based verification \citet{min2023factscore} and self-evaluation \citet{manakul2023selfcheckgpt}. These methods are still expensive to perform even if it's just cluster representatives that we evaluate. Consistency between model responses when used as a metric resulted in significant gains in reasoning. In our case, a measure of consistency is readily available in the form of the strength of the clusters. 
Hence, we use the strength of individual clusters to pick the most consistent of them. Specifically, all clusters having count below a fixed threshold $\Theta$ (tuned on a validation set) are filtered and the consistent clusters' representatives are selected $\{c_1, c_2, c_3,..c_z\}$. Note that we also experimented with self evaluation in our preliminary analysis but found consistency-based measures worked better. We also compare with retrieval-based method to filter clusters in the experiments.

\subsection{Summarizing Selected Clusters}\label{sec:final_answer} 
 The representatives $\{c_1, c_2, c_3,...c_z\}$ are fed into the same LLM $\mathcal{L}$ with a summarizing prompt $P_{combine}$(mentioned in \S \ref{appendix}) to produce the final answer $a_{\asc}$. Note that the number of representatives $z<m$ (most times) and hence is much easier to process by $\mathcal{L}$ compared to \usc\ (feeds $m$ inputs).  This call to the LLM not only summarizes the selected cluster representatives but also filters any slack/filler sentences that were selected by the consistency metric. Overall pipeline is shown in Fig. \ref{fig_method}.\\

\noindent\textbf{Adapting \asc\ to list style datasets}:
We also extend the aforementioned \asc\ methodology to list style datasets. Here, for a given question, the answer produced is typically a list of entities. Following \citet{agrawal2023language}, we directly used each item in the list as an atomic fact.   We simply used surface-form based clustering where two atomic units are placed in the same cluster if their normalized edit distance is below a threshold. $\Theta$ threshold (tuned on a validation set) based filtering is applied to select consistent clusters. The first item in each cluster is considered its representative. The final answer is just the list of selected cluster representatives $a_{\asc} = [c_1, c_2, c_3,...c_z]$.

\section{Experiments}

\subsection{Tasks}
\label{sec:tasks}
\textbf{ASQA} \cite{stelmakh2022asqa}: ASQA is a long-form factoid dataset comprising ambiguous questions. The ambiguous nature of the questions requires answers to comprise diverse facts from multiple documents. The dataset provides individual reference disambiguating short answers for each question and also a reference long answer combining all short answers. Following \citet{gao2023enabling}, performance on this dataset is evaluated by 1. `Str\_EM': checking if reference short answers have an exact match in the LLM generated answer, 2. `QA-F1': Does an external QA model identify these short answers from reference disambiguating questions. Str\_Em is very closely related to the recall of atomic facts relevant to the question. Additionally, we also present the `Mauve' score which compares the fluency and style of the model generated text to the reference answer.

\noindent\textbf{QAMPARI} \cite{rubin2022qampari}: QAMPARI is a list-style factoid QA dataset constructed from Wikipedia knowledge graphs and tables with the questions paraphrased by humans. Performance over this dataset is evaluated by `Precision', `Recall' and `F1' between the generated answer list and reference answer list. As the reference lists are often huge, another measure `Recall-5' scores the answer 100 if at least 5 correct entities are present.

\noindent\textbf{QUEST} \cite{malaviya2023quest}: QUEST is another list-style dataset constructed using Wiki category lists. This is a much more challenging dataset compared to QAMPARI. Following \citet{dhuliawala2023chainofverification}, we transform each category name into a question by prepending "\textit{Name Some}". For Eg. \textit{"Name Some Mary Stewart novels"}. Performance over this dataset is evaluated by Precision, Recall, F1 and Recall-5. 

\noindent\textbf{ELI5} \cite{fan2019eli5}: This is a long-form QA dataset containing how/why/what questions from Reddit. \citet{gao2023enabling} had generated three sub-claims from each golden answer and showed that an answer's entailment score over these sub-claims provides a more accurate measure of its correctness. We use this same `Claim-Recall' to measure the correctness of a generated answer in this work. Similar to Str\_EM in ASQA, this again is very related to the recall of atomic facts relevant for the question. We use the same randomly sampled 1000 questions from the eval set as from \citet{gao2023enabling}.
We use the test sets for QAMPARI, QUEST and validation sets for ASQA, ELI5.

\subsection{Methods} 
We compare the following methods.\\
\textbf{\asc}: The method that we propose in this work. Splits multiple samples into smaller parts, and clusters them. The best clusters (picked by consistency score) are summarised using an LLM.\\
\textbf{\fc}: Abbrevated from Factual correctness. Similar to \ours\ but uses FactScore \cite{min2023factscore} to pick the relevant clusters instead of the cluster strength in \S \ref{c_c}. We use 5 retrieved passages and InstructGPT from FactScore to verify the correctness of each cluster. \fc\ doesn't use any consistency measures despite sharing the name with \asc.\\
\textbf{\usc}: Consistency based method proposed in \citet{chen2023universal}. \usc\ prompts an LLM with a list of answers and asks it to pick the most consistent answer. With a large enough list (m=50 in our case), LLMs tend to underperform \cite{qin2023large}. Hence we use a different method to approximate consistency. We use the same clustering pipeline as mentioned for \asc. Each of the $m$ generations is given a score equal to the sum of the strengths of all clusters it contributes to. The highest scoring generation is selected as the final answer.\\
\textbf{\uscl}: This is the original formulation which used a list of samples as input to the LLM and found the most consistent response. We reduced the input list whenever it did not fit the context window. We observed that this method picks the first response a majority of the time possibly because of longer list problems in LLMs \cite{qin2023large}.  \\
\textbf{\acf}: Abbreviated from Atomic Consistency-based Filter. Only one generation(very first seed) out of the 50 is used. We use cluster strengths from the 50 generations to filter out facts from the one generation. We use the same $\Theta$ used in \asc\ to filter atomic facts. The selected facts are combined into a single answer using a summarization in \S \ref{sec:final_answer}.\\
\textbf{\fcf}: Abbreviated from Factual Correctness based Filter. Similar to \acf, uses a FactScore-based filter to throw out facts from the one generation. Leftover facts are combined as mentioned in \S \ref{sec:final_answer}.\\
\textbf{\direct}: Direct generation from the LLM. Results are averaged over five different seeds.

While Self-Evaluation style methods were also shown to improve the correctness of LLM generations, performing self-evaluation over $m*k$ number of clusters ($m*k$ LLM calls)(similar to \asc) or $m$ number of generations ($m$ LLM calls)(similar to \usc), to answer one single question was very expensive. Note that \ours\ only requires two calls to the LLM - one for generating $m$ answers and one for merging the selected clusters. Clustering in \ours\ is performed by off-the-shelf models.

\begin{table*}[t]
\centering
\resizebox{\textwidth}{!}{\begin{tabular}{l|l||c|c|c|c|c||c|c|c|c}
\toprule
& & \multicolumn{4}{c}{\textit{\quad ASQA}}& & \multicolumn{4}{c}{\textit{ELI5}}\\
\hline
                         &        & \textbf{\#Clusters}    & \textbf{length} & \textbf{Mauve} & \textbf{Str\_EM} & \textbf{QA-F1} &  \textbf{\#Clus.} & \textbf{length}     & \textbf{Mauve}         & \textbf{Claims\_Nli}                   \\\hline
\multirow{7}{*}{ChatGPT} & \direct & \multirow{5}{*}{-}     & 56.29                              & 44.64                             & 37.13 & 29.33 & \multirow{5}{*}{-}     & 104.35 & 24.57 & 18.66  \\
                         & \acf    &                        & 42.99                               & 53.66                              & 36.16 & 28.98 &  & 84.11	& 20.73 & 18.2  \\
                         & \fcf    &                        & 45                                  & 52.68                              & 36.84 & 29.64 & & 94.75   & 27.97   & 18.7  \\
                         & \uscl &  & 56.72 & 44.88 & 37.91 & 29.71&  & 104.13 & 21.11 & 18.76 \\
                         & \usc    &                        & 64.52                               & 40.19                              & 39.05 & 30.88&  &97.36  & 24.09 & 17.4 \\
                         \cline{2-11}
                         & \fc\ (Ours)    &  30.74 & 106.7 & 41.25 & \textbf{44.96} & \underline{31.91}&  56.83	& 172.66 & 22.68 & \textbf{22.16}       \\
                         & \asc\ (Ours)    &     15.7    & 101.17                              & 47.01                              & \underline{44.1} & \textbf{32.22}&  16.68     & 163.58   & 21.29  & \underline{21.43} \\\midrule
\multirow{6}{*}{Llama2}   & \direct & \multirow{4}{*}{-}      & 41.88                              & 68                                 & 28.71 & 23.58 & \multirow{4}{*}{\textbf{-}} & 84.38  & 46.59 & 13.98 \\
                         & \acf    &                        & 25.78                               & 63.79                              & 28.48 & 24.73 &                             & 58.20  & 38.22 & 13.70 \\
                         & \fcf    &                        & 28.71                               & 68.22                              & 28.38 & 24.64 &                             & 66.96  & 35.20 & 14.57 \\
                         & \usc    &                        & 63.7                                & 63.63                              & \underline{33.16} & \underline{26.42}&                             & 115.82 & 35.21 & 17.70 \\\cline{2-11}
                         & \fc\ (Ours)     & 33.57 & 108.18 & 62.68 & \textbf{39.26} & \underline{26.54} & 83.42   & 148.30 & 35.25 & \underline{18.97} \\
                         & \asc\ (Ours)    &     12.68     & 91.91                               & 70.52                              & \underline{38.82} & \textbf{27.16}& 14.32   & 143.07 & 28.09 & \textbf{19.40}\\\bottomrule
\end{tabular}}
\caption{ASQA, ELI5 results. \asc\ does the best on QA-F1 and demonstrates strong Str\_EM. \fc\ picks a large number of clusters and does well on Str\_EM. \asc\ also demonstrates strong Mauve. \asc, \fc\ achieve best Claims\_Nli score on ELI5. Results justify that merging of samples is better than picking one sample.}
\label{tab:asqa_results}
\end{table*}

\begin{table*}[ht]
\centering
\resizebox{\textwidth}{!}{\begin{tabular}{l|l||c|c|c|c|c|c||c|c|c|c|c|c}
\toprule
&&\multicolumn{5}{c}{\textit{QAMPARI}} && \multicolumn{6}{c}{\textit{QUEST}}                                                                                                                                                                                                                     \\\hline
                         & \textbf{Method} &\textbf{\#Pred} & \textbf{Prec} & \textbf{Rec} & \textbf{Rec-5} & \textbf{F1} & \textbf{F1-5}&\textbf{\#Pred} & \textbf{Prec} & \textbf{Rec} & \textbf{Rec-5} & \textbf{F1} & \textbf{F1-5} \\\hline
\multirow{7}{*}{ChatGPT} & \direct & 5.2   & 21.35 & 13.82 & 23.47 & 15.35 & 21.83 & 5.56  & 12.05 & 6.76 & 12.91 & 7.45  & 11.6  \\
                         & \acf    & 3.61  & 24.16 & 12.5  & 21.96 & 15.04  & 22.18  & 3.07   & 14.71 & 5.65  & 10.67  & 7.06   & 11.53 \\
                         & \fcf   & 4.41  & 22.59 & 13.29 & 23.16 & 15.33 & 22.16  & 3.61   &	13.55 &	5.91  &	11.03  & 7.01   & 11.27 \\
                         & \uscl    & 4.95 & 20.88 & 13.39 & 22.91 & 14.94 & 21.33  & 5.10   & 11.86 & 6.18  & 11.92  & 7.08   & 11.16 \\
                         & \usc    & 8.97  & 20.7  & 19.21 & 31.28 & \underline{18.07}  & \underline{24.2}& 7.83   & 11.98 & 8.43  & 15.19  & 8.23   & \underline{12.21} \\\cline{2-14}
                         & \fc\    & 40.83 & 13.42 & 29.81 & 45.04 & 15.7 & 18.82  & 39.9 & 7.94 & 17.31 & 30.73 & \underline{8.47} & 10.84 \\
                         & \asc\    & 7.09  & 22.98 & 20.5  & 33.04 & \textbf{19.46}  & \textbf{26.21}& 8.44   & 12.47 & 10.41 & 19.15  & \textbf{9.75}   & \textbf{14.09} \\\midrule
\multirow{6}{*}{Llama2}  & \direct & 4.86 & 13.5  & 9.25  & 16.23 & 10.22  & 14.47& 5.46 & 6.74  & 4.16  & 7.66  & 4.42 & 6.7   \\
                         & \acf    & 3.17  & 14.94 & 7.96  & 13.84 & 9.69   & 13.85  & 3.48   & 7.9   & 3.47  & 6.34   & 4.14   & 6.54  \\
                         & \fcf    & 3.88  & 14.1  & 8.93  & 15.36 & 10.15  & 14.22  & 3.43   & 8.06  &	3.78  &	6.75   & 4.38   & 6.77 \\
                         & \usc   & 7.44  & 14.07 & 11.61 & 20.04 & \underline{11.64}  & \underline{15.99} & 9.36   & 7.76  & 5.4   & 10.16  & \underline{5.38}   & \textbf{7.96}  \\\cline{2-14}
                         & \fc\   & 27.35 & 10.74 & 18.44 & 29.88 & 11.52 & 14.4 & 28.07 & 5.63 & 10.64 & 19.08 & \textbf{5.81} & 7.67\\
                         & \asc\   & 6.08  & 14.51 & 12.15 & 20.58 & \textbf{12.15} & \textbf{16.44} & 6.77   & 7.42  & 5.52  & 9.97   & 5.3    & \underline{7.86} \\\bottomrule                         
\end{tabular}}
\caption{\asc\ outperforms \direct, \usc\ and \fc. \fc\ picks a large number of clusters and does worse on P, F1, F1-5. Results justify that consistency-based cluster selection does better than retrieval-based cluster selection.}
\label{tab:qampari_results}
\vspace{-1.1em}
\end{table*}

\subsection{Model Details and Setup}
We use $m=50$ for generations. The same set of generations are used by \asc, \fc, \usc. \direct\ is the average of five seeds among the 50. \acf, \fcf\ use only one seed for the answer. \acf\ uses all 50 seeds but for building a consistency-based selection mechanism. Sentence tokenization was performed by NLTK \cite{bird2009natural}. Sentence embeddings for clustering were generated using robert-large SimCSE \cite{gao2021simcse}, agglomerative clustering ($d=0.15$) is used to perform clustering in ASQA, ELI5. For \fc, \fcf, retrieval, query and document embeddings were generated using \textsc{gtr-t5-xxl} \cite{ni2021large} and Wikipedia following \cite{gao2023enabling}. In \fc, \fcf, each cluster/fact used 5 retrieved passages for verification. InstructLlama model from \citet{min2023factscore} was used to verify facts in \fc,\fcf. A fact was called true if at least 1 of the 5 passages supported it. Experiments on all four datasets were performed with both ChatGPT, Llama-2(70b) \cite{touvron2023llama}. In all the experiments, the same LLM is used to perform both generation and summarization. Generation and summarization prompts along with other details are presented in \S \ref{ap:implement_details}.

\asc\ uses hyperparam $\Theta$ tuned over the development set to maximise F1-5 for QAMPARI and QUEST. For ASQA, ELI5, $\Theta$ was chosen such that the number of selected clusters comfortably fit the context window of ChatGPT, Llama2. \acf\ used the same threshold as \asc\ in filtering atomic facts. \fcf, \usc\ did not require tuning any hyperparameters. Clustering parameters were same for all models that used it. More details in \S \ref{ap:implement_details}

\begin{table*}[t]
\centering
\resizebox{\textwidth}{!}{
\begin{tabular}{c|l||c|c|c|c|c||c|c|c|c|c|c}
\toprule
&&\multicolumn{4}{c}{\textit{ASQA}}&&\multicolumn{6}{c}{\textit{QAMPARI}}\\\hline
\textbf{Ablation} & \textbf{Method}                & \textbf{\#Clusters} & \textbf{length} & \textbf{Mauve} & \textbf{Str\_EM} & \textbf{QA-F1}& \textbf{\#Pred} & \textbf{Prec} & \textbf{Rec} & \textbf{Rec-5} & \textbf{F1} & \multicolumn{1}{l}{\textbf{F1-5}} \\\hline
\multirow{3}{*}{1} & \asc\                            & 15.7                                    & 101.17                              & 47.01                              & \textbf{44.1}                                 & \textbf{32.22}& 7.09                                        & 22.98                             & 20.5                             & 33.04                              & \textbf{19.46}                           & \textbf{26.21}                             \\
& \rc              & 15.7                                    & 85.31                               & 49.97                              & \underline{42.62}                                & \underline{31.75}& 7.09                                        & 11.86                             & 10.08                            & 18.62                              & 9.77                            & 14.05                             \\
& \rs & 15.7                                    & 99.45                               & 42.08                              & 41.5                                 & 29.36& 7.09                                        & 22.19                             & 13.8                             &    24.42                        & 15.39                           & 22.1                              \\\midrule
\multirow{3}{*}{2} & \usc    &          \multirow{2}{*}{-}         & 64.52                               & 40.19                              & 39.05                                & 30.88 & 8.97  & 20.7  & 19.21 & 31.28 & \underline{18.07}  & \underline{24.2} \\
& \texttt{High Token/\#Pred}             &                                       & 82.93                               & 40.59                              & 37.8                                 & 28.79& 10.48                                       & 17.19                             & 18.3                             & 29.28                              & 16.07                           & 21.01                            \\\bottomrule
\end{tabular}}
\caption{Ablation 1: ASC performs better than randomly picking clusters and randomly picking sentences on ASQA, QAMPARI. Ablation 2: Larger length or higher \#Predictions in response is not critical for better performance.}
\label{tab:asqa_ab1}
\end{table*}

\subsection{Results}
\noindent Table \ref{tab:asqa_results} demonstrates the eval set results for ASQA. \asc\ and \fc\ do much better than \usc\ in Str\_EM (exact match) and QA-F1 (QA model performance). This shows that merging parts of multiple answers performs better than picking a single answer. \acf\ and \fcf\ are more picky in selecting what facts can show up in the final answer and hence have worse STR\_EM, QA-F1 compared to \direct. The short answers in \acf, \fcf\ help achieve a better Mauve score. \asc\ beats \direct, \usc, and \fc\ in achieving better Mauve score. 
Table \ref{tab:asqa_results} also shows the eval set results of ELI5. \asc, \fc\ perform better than all other models yet again demonstrating the strength of merging multiple model responses. Mauve is lower in this case because the reference answers are from Reddit subposts and the style didn't match the long answers generated by \asc. As will be shown in ablations, \asc\ offers easy control over Mauve by changing $\Theta$. Fig. \ref{fig:eli5_vary} shows that $\Theta$ can be adjusted to improve \asc's Mauve score over other methods while still retaining Claims\_Nli.\\ Table \ref{tab:qampari_results} shows test set results for QAMPARI. \asc\ performs the best. \fc\ similar to the previous case selects a large number of clusters. It does well on recall but significantly drops on precision leading to a worse overall performance. This also shows that longer answers are not always helpful. Since this is a list-style dataset, we also show \#Predictions (size of the list) which is somewhat equivalent to \#Clusters in the previous case. Note that \asc\ beats \usc\ despite having lower \#Predictions. As expected, \acf, \fcf\ have much lower \#Predictions and have higher precision compared to others. \asc\ relies on consistency to predict a much higher \#Predictions while also matching/improving precision. This strongly justifies the strength of consistency-based cluster selection. Specifically, it achieves the two goals we had set out in this exploration. It removes incorrect atomic facts from a sample (increase in precision compared to $\usc$) and adds correct atomic facts from other samples (increase in recall compared to $\usc$).
Table \ref{tab:qampari_results} also shows test set results for QUEST. The trends are very similar to QAMPARI with \asc\ performing the best using ChatGPT. With Llama2, \asc\, \fc\, and \usc\ all show similar improvements over the \direct\ baseline.
The basic ideas of merging and consistency still show promise,
but QUEST is a difficult dataset with compositional logic where zeroshot Llama2 generally performs poorly.
Summarizing all the observations from above, we conclude that\\ 1. Merging LLM samples is better than selecting one single sample.\\ 2. Atomic-Consistency is a strong measure to select clusters and improve correctness.


\subsection{Ablations}
\subsubsection{Ablation 1: Dissecting ASC}
To effectively understand the contribution of different components of ASC, we analyze the effect of each subcomponent. Table \ref{tab:asqa_ab1} shows results with ASQA. ASC first clusters individual sentences from all 50 generations and merges sentences with high cluster strength using an LLM. The \rc\ method follows the same clustering as \asc\ but randomly picks clusters before merging them using an LLM. \rs\ doesn't perform any clustering and randomly picks the sentences from all generations and summarizes them. In both runs, we pick the exact same number of sentences that were picked by \asc. \rc\ drops both Str\_Em and QA-F1 but still does better than \direct, \usc\ in Table \ref{tab:asqa_ab1}. This shows the strength of including diverse sentences from multiple samples into the answer generation. In the case of ASQA and ELI5, there is less hallucination compared to QAMPARI and QUEST. Hence, randomly picking clusters does fairly well better than most other baselines. \rs\ further drops in metrics while still maintaining a high Str\_EM. 

Table \ref{tab:asqa_ab1} also does the same analysis for QAMPARI. Here again, we run the two random baselines. Similar to the previous case, \asc\ does the best.

\subsubsection{Ablation 2: Longer length answers}
 From Table \ref{tab:qampari_results}, \ref{tab:asqa_results}, \asc\ and \fc\ often have higher length answers and higher \#predictions. One might deduce that longer generations tend to give better results on the datasets tested. Hence, we perform an additional experiment which picks the longest length sample (among the 50 samples) for ASQA and pick the sample with the highest \#Predictions (among the 50 samples) for QAMPARI as the final answer. Results are shown in Table \ref{tab:asqa_ab1}. Despite having a larger length or higher \#predictions, \usc\ with a lower length and lower \#predictions perform much better. This shows length is not the most important factor for improved performance.

\subsubsection{Ablation 3: ASC is simple to control with \texorpdfstring{$\Theta$}{Theta} (Sensitivity Analysis)}
\begin{figure}[ht]     
     \centering
     \includegraphics[width=\columnwidth]{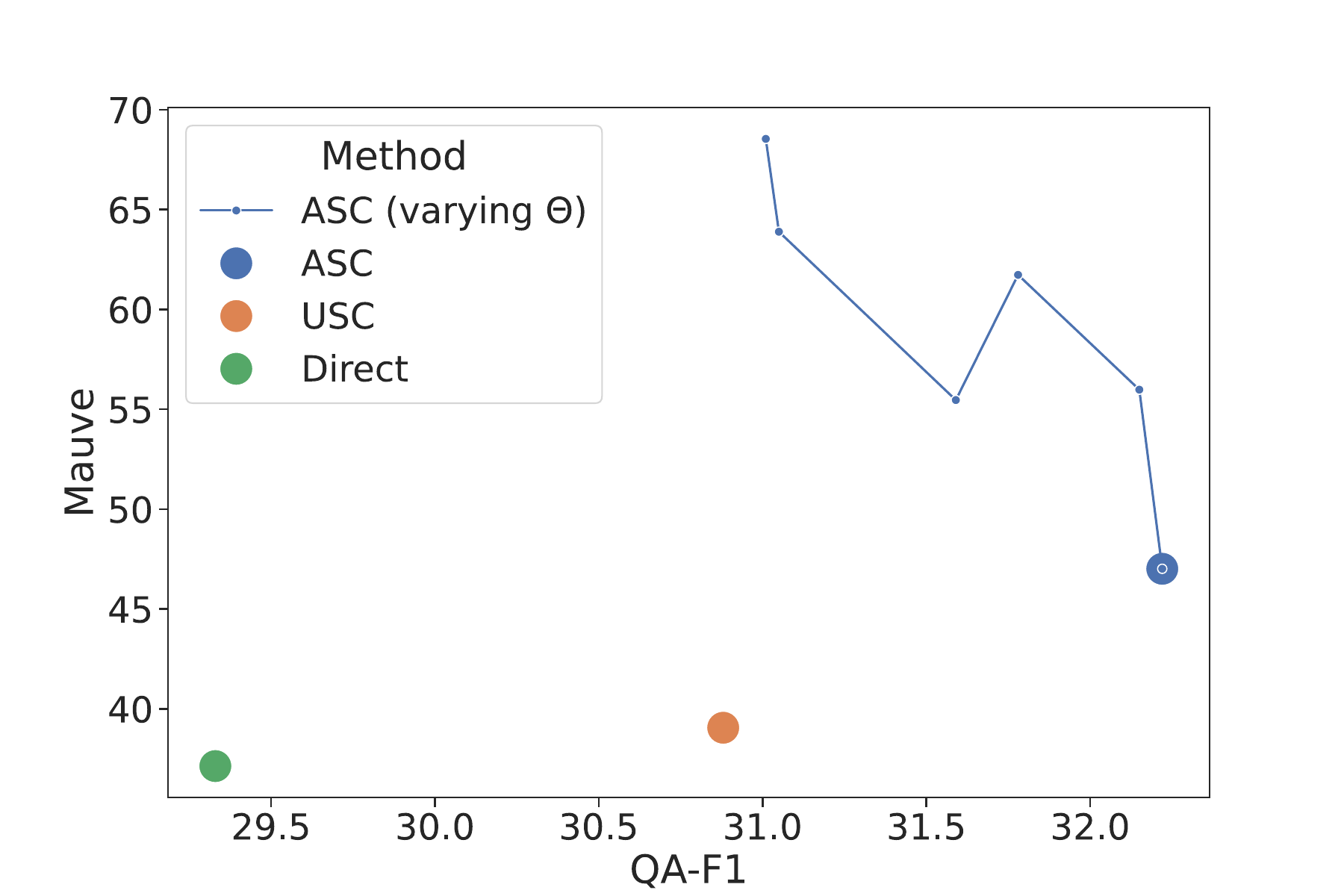}
    \caption{ASQA. Increasing $\Theta$ improves QA-F1, reduces Mauve. Adjusting $\Theta$ produces a preferred answer.}
    \label{fig:asqa_vary}
    \vspace{-1em}
\end{figure}

$\Theta$ is a parameter which critically affects the performance of \asc.
For Table \ref{tab:qampari_results},  we used a value of $\Theta$ that performed the best on the validation set. Different number of selected clusters result in differences in various performance metrics. For example, \fc\ which selects a large number of clusters is more suited to high recall scenarios where precision is less important. Hence, to better understand this effect, we experiment with different values of $\Theta$ in this subsection. 
Fig. \ref{fig:asqa_vary} shows the effect of varying $\Theta$ on ASQA.  A lower $\Theta$ resulted in selecting a large number of clusters and resulted in improving QA-F1. This also increased the length of the final response. Reducing $\Theta$ on the other hand improved the Mauve fluency score as the shorter final answer matched more with the reference answer. Hence, one might easily adjust $\Theta$ to obtain an answer aligned with their preference (Mauve or QA-F1). From the Fig. \ref{fig:asqa_vary}, \asc\ can outperform all other methods in Mauve(can achieve >65) while still retaining a QA-F1 (>31). The best of other methods was Mauve (53.66) and QA-F1 (30.99). A similar result was seen in ELI5 where increasing $\Theta$ achieved the highest Mauve \S \ref{fig:eli5_vary}. 

Fig. \ref{fig:qampari_vary} in \S \ref{appendix} shows the effect of varying $\Theta$ on QAMPARI. The relationship here is more linear. Increasing $\Theta$ results in fewer clusters with high strength (high  precision). Reducing $\Theta$ results in higher recall. \fc's criterion enabled it to select a larger number of clusters resulting in higher recall. Here again, one can easily change $\Theta$ to obtain an answer with preferred qualities. 

\begin{figure}     
     \centering
     \includegraphics[width=\columnwidth]{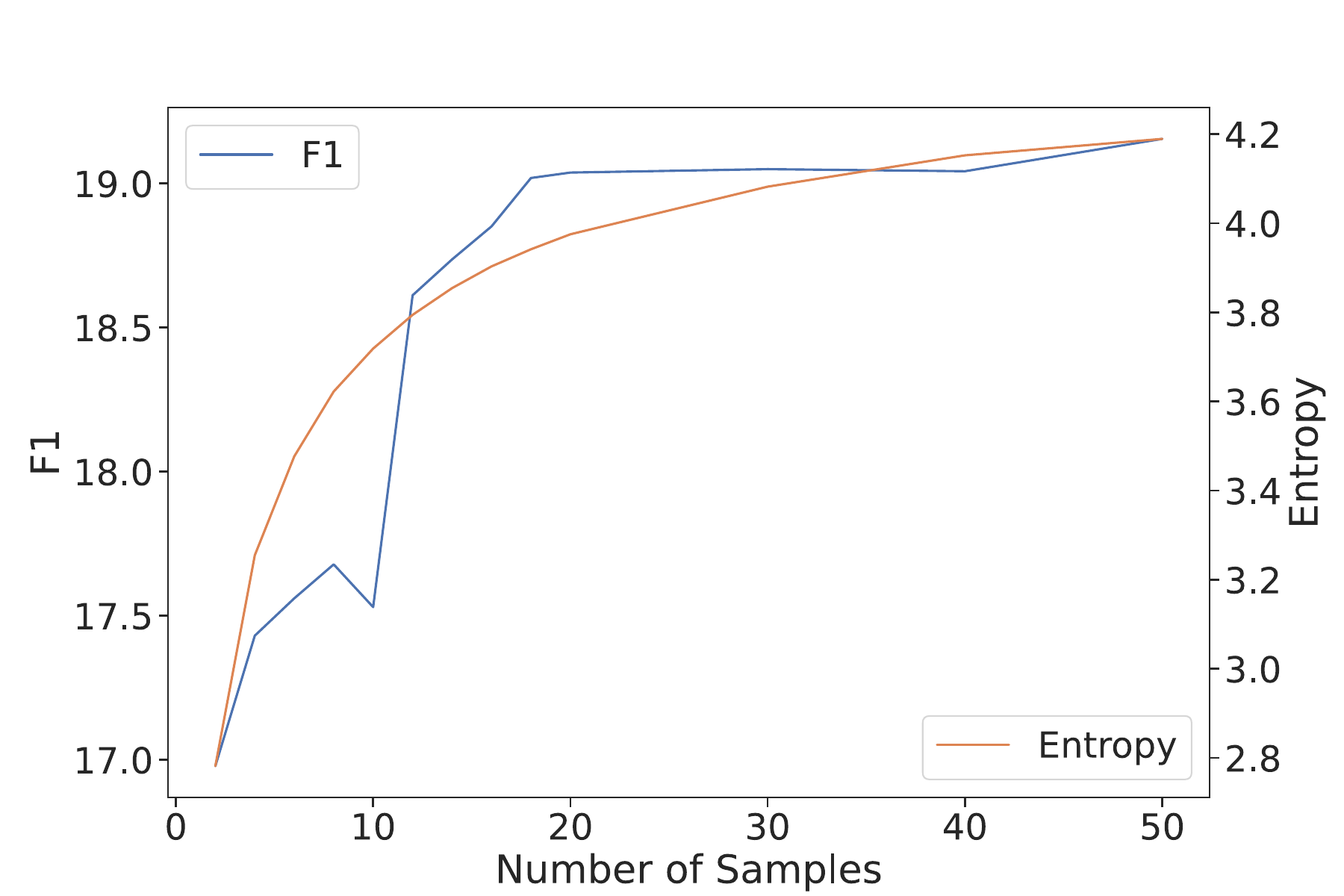}
    \caption{QAMPARI. Performance starts to stagnate when clusters' entropy stagnates.}
    \label{fig:entropy}
    \vspace{-2em}
\end{figure}

\subsection{Analysis: Can ASC work with fewer number of generations? Use Entropy}
Cost of generation using an LLM linearly scales with the number of samples. ASC used a large number of samples, $m=50$ in our earlier experiments. It might not always be feasible to generate this large number of samples due to time and budget constraints. In this subsection, we investigated if we could generate fewer samples and yet capture the gains provided by \asc. While we focused on QAMPARI for this analysis, we found similar trends with other datasets as well. We first looked at how the entropy of the clusters (considering each cluster to have a probability proportional to its strength) changes with increasing number of generations. In the beginning ($m=1$), all clusters have one member and equal probability. Hence, the entropy is lower. As and when more samples get added, some clusters accumulate higher strength and some others remain low strength. Hence, the entropy increases due to unequal probabilities of clusters. We empirically found that entropy starts to stagnate with higher values of $m$. To measure the performance of $m$ samples, we scaled the optimal $\Theta$ we found in Table. \ref{tab:qampari_results} by $\frac{m}{50}$. We found that performance follows a similar trend increasing quickly at the beginning while slowly stagnating. Performance and Entropy curves values with $m$ are shown in Fig. \ref{fig:entropy}. Interestingly, performance starts to stagnate right around when entropy starts to stagnate. Entropy stagnation can thus be used as an indication to stop generating more samples from the LLM and fix $m$.

\subsection{Analysis: Room for improvement}\label{ss:o_recall}

\begin{table}[ht]
\centering
\resizebox{\columnwidth}{!}{\begin{tabular}{l|c||c|c||c|c}
\toprule
&&\multicolumn{1}{r}{\textit{ASQA}}&&\multicolumn{2}{c}{\textit{QAMPARI}}\\
\hline
\textbf{Method}         & \textbf{\#Gen} & \textbf{Str\_EM} & \textbf{QA-F1}& \textbf{Rec} & \textbf{Rec-5} \\ \hline
\multirow{6}{*}{Oracle} & 1                      & 36.32            & 22.88 & 13.94                            & 24.24                              \\
                        & 2                      & 40.64            & 28.05& 18.15                            & 30.46                              \\
                        & 5                      & 45.65            & 34.03& 24.53                            & 39.02                              \\
                        & 15                     & 50.97            & 39.28& 32.29                            & 48.78                              \\
                        & 25                     & 53.1             & 41.29& 35.86                            & 52.76                              \\
                        & 50                     & 56.09            & 45.2& 40.06                            & 56.90                              \\\hline
ASC                     & 50                     & 44.1             & 32.22& 20.50                            & 33.04                              \\ \bottomrule
\end{tabular}}
\caption{Oracle results reveal sizable scope for improvement using our approach of merging multiple responses.}
\label{tab:asqa_oracle}
\end{table}

To better understand the gains of \asc, we look at the best possible performance offered by our mechanism of merging multiple sample generations. We use the same 50 generations that were used to produce the \asc\ results. Table \ref{tab:asqa_oracle} shows the best possible performance (Oracle) with the number of generations. Exact procedure for oracle numbers in \S \ref{ap:orn}.

The experiment presents interesting observations. 1. Using just five generations significantly increases the oracle performance. 2. Oracle's performance stagnates at a higher number of generations. Our observations on \asc\ performance stagnating after 20 generations are in line with these results. 3. \asc\ only captures $20.50$ of the $40.06$ possible recall on QAMPARI and $44.1$ of the $56.09$ possible Str\_EM on ASQA. Thus while \asc\ captures a fair share of the performance gain offered by merging multiple generations, a sizable portion of performance gain still remains untapped. Future work aims at capturing this potential gain by using stronger verification methods involving a combination of \asc\ with \fc\ and methods in \S \ref{sec:rel_work}.

\section{Conclusion}
In this work, we propose \asc, a simple way of merging subparts of multiple answer samples produced by an LLM. Through extensive experiments and ablations, we show the 1. Benefits of merging subparts of multiple answers over picking one single answer. 2. Strength of \textit{consistency} as a measure for improving correctness. \asc\ significantly outperforms \usc, a strong baseline for generating long-form answers. We show empirical evidence for minimizing the number of samples required by \asc. Finally, our analysis also reveals untapped potential for enhancing long-form generations using our approach of merging multiple responses.

\section{Limitations}
\textbf{Other verification methods not tried}
Our approach of merging consistent clusters requires $m*k$ number of LLM calls per cluster. It was beyond our budget limitations to try methods like Self-Evaluation\cite{ren2023self} or Self-Verification \cite{dhuliawala2023chainofverification}. That said, for future work we intend to use a combination of consistency and these other measures. One could just use consistency for high strength atomic clusters and self-verification/self-eval for lower frequency clusters which cannot be judged by consistency.\\
\noindent\textbf{Smaller language models not tried}
Some of the datasets used in our work are very challenging and not suitable for smaller language models. To effectively prove the strength of our approach, we stuck to ChatGPT, Llama-70b.\\\\\\\\\\

\noindent\textbf{Broader Impact and Discussion of Ethics}:\\
While our model is not tied to any specific applications, it could be used in
sensitive contexts such as health-care, etc.  Any work using our
method is requested to undertake extensive quality-assurance and
robustness testing before applying in their setting. To the best of our knowledge, the datasets used in our work do not contain any sensitive information.\\

\noindent\textbf{Replicability}:\\
Code: \href{https://github.com/raghavlite/ASC}{https://github.com/raghavlite/ASC}

\bibliography{custom}

\begin{thebibliography}{26}
\expandafter\ifx\csname natexlab\endcsname\relax\def\natexlab#1{#1}\fi

\bibitem[{Agrawal et~al.(2023)Agrawal, Mackey, and Kalai}]{agrawal2023language}
Ayush Agrawal, Lester Mackey, and Adam~Tauman Kalai. 2023.
\newblock Do language models know when they're hallucinating references?
\newblock \emph{arXiv preprint arXiv:2305.18248}.

\bibitem[{Ahn et~al.(2022)Ahn, Brohan, Brown, Chebotar, Cortes, David, Finn, Fu, Gopalakrishnan, Hausman et~al.}]{ahn2022can}
Michael Ahn, Anthony Brohan, Noah Brown, Yevgen Chebotar, Omar Cortes, Byron David, Chelsea Finn, Chuyuan Fu, Keerthana Gopalakrishnan, Karol Hausman, et~al. 2022.
\newblock Do as i can, not as i say: Grounding language in robotic affordances.
\newblock \emph{arXiv preprint arXiv:2204.01691}.

\bibitem[{Arslan et~al.(2020)Arslan, Hassan, Li, and Tremayne}]{arslan2020benchmark}
Fatma Arslan, Naeemul Hassan, Chengkai Li, and Mark Tremayne. 2020.
\newblock A benchmark dataset of check-worthy factual claims.
\newblock In \emph{Proceedings of the International AAAI Conference on Web and Social Media}, volume~14, pages 821--829.

\bibitem[{Bird et~al.(2009)Bird, Klein, and Loper}]{bird2009natural}
Steven Bird, Ewan Klein, and Edward Loper. 2009.
\newblock \emph{Natural language processing with Python: analyzing text with the natural language toolkit}.
\newblock " O'Reilly Media, Inc.".

\bibitem[{Bubeck et~al.(2023)Bubeck, Chandrasekaran, Eldan, Gehrke, Horvitz, Kamar, Lee, Lee, Li, Lundberg et~al.}]{bubeck2023sparks}
S{\'e}bastien Bubeck, Varun Chandrasekaran, Ronen Eldan, Johannes Gehrke, Eric Horvitz, Ece Kamar, Peter Lee, Yin~Tat Lee, Yuanzhi Li, Scott Lundberg, et~al. 2023.
\newblock Sparks of artificial general intelligence: Early experiments with gpt-4.
\newblock \emph{arXiv preprint arXiv:2303.12712}.

\bibitem[{Chen et~al.(2023)Chen, Aksitov, Alon, Ren, Xiao, Yin, Prakash, Sutton, Wang, and Zhou}]{chen2023universal}
Xinyun Chen, Renat Aksitov, Uri Alon, Jie Ren, Kefan Xiao, Pengcheng Yin, Sushant Prakash, Charles Sutton, Xuezhi Wang, and Denny Zhou. 2023.
\newblock Universal self-consistency for large language model generation.
\newblock \emph{arXiv preprint arXiv:2311.17311}.

\bibitem[{Dhuliawala et~al.(2023)Dhuliawala, Komeili, Xu, Raileanu, Li, Celikyilmaz, and Weston}]{dhuliawala2023chainofverification}
Shehzaad Dhuliawala, Mojtaba Komeili, Jing Xu, Roberta Raileanu, Xian Li, Asli Celikyilmaz, and Jason Weston. 2023.
\newblock \href {http://arxiv.org/abs/2309.11495} {Chain-of-verification reduces hallucination in large language models}.

\bibitem[{Elaraby et~al.(2023)Elaraby, Lu, Dunn, Zhang, Wang, and Liu}]{elaraby2023halo}
Mohamed Elaraby, Mengyin Lu, Jacob Dunn, Xueying Zhang, Yu~Wang, and Shizhu Liu. 2023.
\newblock Halo: Estimation and reduction of hallucinations in open-source weak large language models.
\newblock \emph{arXiv preprint arXiv:2308.11764}.

\bibitem[{Fan et~al.(2019)Fan, Jernite, Perez, Grangier, Weston, and Auli}]{fan2019eli5}
Angela Fan, Yacine Jernite, Ethan Perez, David Grangier, Jason Weston, and Michael Auli. 2019.
\newblock Eli5: Long form question answering.
\newblock \emph{arXiv preprint arXiv:1907.09190}.

\bibitem[{Gao et~al.(2021)Gao, Yao, and Chen}]{gao2021simcse}
Tianyu Gao, Xingcheng Yao, and Danqi Chen. 2021.
\newblock Simcse: Simple contrastive learning of sentence embeddings.
\newblock \emph{arXiv preprint arXiv:2104.08821}.

\bibitem[{Gao et~al.(2023)Gao, Yen, Yu, and Chen}]{gao2023enabling}
Tianyu Gao, Howard Yen, Jiatong Yu, and Danqi Chen. 2023.
\newblock Enabling large language models to generate text with citations.
\newblock \emph{arXiv preprint arXiv:2305.14627}.

\bibitem[{Liu et~al.(2023)Liu, Zhang, and Liang}]{liu2023evaluating}
Nelson~F Liu, Tianyi Zhang, and Percy Liang. 2023.
\newblock Evaluating verifiability in generative search engines.
\newblock \emph{arXiv preprint arXiv:2304.09848}.

\bibitem[{Malaviya et~al.(2023)Malaviya, Shaw, Chang, Lee, and Toutanova}]{malaviya2023quest}
Chaitanya Malaviya, Peter Shaw, Ming-Wei Chang, Kenton Lee, and Kristina Toutanova. 2023.
\newblock Quest: A retrieval dataset of entity-seeking queries with implicit set operations.
\newblock \emph{arXiv preprint arXiv:2305.11694}.

\bibitem[{Manakul et~al.(2023)Manakul, Liusie, and Gales}]{manakul2023selfcheckgpt}
Potsawee Manakul, Adian Liusie, and Mark~JF Gales. 2023.
\newblock Selfcheckgpt: Zero-resource black-box hallucination detection for generative large language models.
\newblock \emph{arXiv preprint arXiv:2303.08896}.

\bibitem[{Min et~al.(2023)Min, Krishna, Lyu, Lewis, Yih, Koh, Iyyer, Zettlemoyer, and Hajishirzi}]{min2023factscore}
Sewon Min, Kalpesh Krishna, Xinxi Lyu, Mike Lewis, Wen-tau Yih, Pang~Wei Koh, Mohit Iyyer, Luke Zettlemoyer, and Hannaneh Hajishirzi. 2023.
\newblock Factscore: Fine-grained atomic evaluation of factual precision in long form text generation.
\newblock \emph{arXiv preprint arXiv:2305.14251}.

\bibitem[{Ni et~al.(2021)Ni, Qu, Lu, Dai, {\'A}brego, Ma, Zhao, Luan, Hall, Chang et~al.}]{ni2021large}
Jianmo Ni, Chen Qu, Jing Lu, Zhuyun Dai, Gustavo~Hern{\'a}ndez {\'A}brego, Ji~Ma, Vincent~Y Zhao, Yi~Luan, Keith~B Hall, Ming-Wei Chang, et~al. 2021.
\newblock Large dual encoders are generalizable retrievers.
\newblock \emph{arXiv preprint arXiv:2112.07899}.

\bibitem[{Pedregosa et~al.(2011)Pedregosa, Varoquaux, Gramfort, Michel, Thirion, Grisel, Blondel, Prettenhofer, Weiss, Dubourg et~al.}]{pedregosa2011scikit}
Fabian Pedregosa, Ga{\"e}l Varoquaux, Alexandre Gramfort, Vincent Michel, Bertrand Thirion, Olivier Grisel, Mathieu Blondel, Peter Prettenhofer, Ron Weiss, Vincent Dubourg, et~al. 2011.
\newblock Scikit-learn: Machine learning in python.
\newblock \emph{Journal of machine learning research}, 12(Oct):2825--2830.

\bibitem[{Qin et~al.(2023)Qin, Jagerman, Hui, Zhuang, Wu, Shen, Liu, Liu, Metzler, Wang et~al.}]{qin2023large}
Zhen Qin, Rolf Jagerman, Kai Hui, Honglei Zhuang, Junru Wu, Jiaming Shen, Tianqi Liu, Jialu Liu, Donald Metzler, Xuanhui Wang, et~al. 2023.
\newblock Large language models are effective text rankers with pairwise ranking prompting.
\newblock \emph{arXiv preprint arXiv:2306.17563}.

\bibitem[{Ren et~al.(2023)Ren, Zhao, Vu, Liu, and Lakshminarayanan}]{ren2023self}
Jie Ren, Yao Zhao, Tu~Vu, Peter~J Liu, and Balaji Lakshminarayanan. 2023.
\newblock Self-evaluation improves selective generation in large language models.
\newblock \emph{arXiv preprint arXiv:2312.09300}.

\bibitem[{Rubin et~al.(2022)Rubin, Yoran, Wolfson, Herzig, and Berant}]{rubin2022qampari}
Samuel Joseph Amouyal~Ohad Rubin, Ori Yoran, Tomer Wolfson, Jonathan Herzig, and Jonathan Berant. 2022.
\newblock Qampari:: An open-domain question answering benchmark for questions with many answers from multiple paragraphs.
\newblock \emph{arXiv preprint arXiv:2205.12665}.

\bibitem[{Shi et~al.(2022)Shi, Fried, Ghazvininejad, Zettlemoyer, and Wang}]{shi2022natural}
Freda Shi, Daniel Fried, Marjan Ghazvininejad, Luke Zettlemoyer, and Sida~I Wang. 2022.
\newblock Natural language to code translation with execution.
\newblock \emph{arXiv preprint arXiv:2204.11454}.

\bibitem[{Stelmakh et~al.(2022)Stelmakh, Luan, Dhingra, and Chang}]{stelmakh2022asqa}
Ivan Stelmakh, Yi~Luan, Bhuwan Dhingra, and Ming-Wei Chang. 2022.
\newblock Asqa: Factoid questions meet long-form answers.
\newblock \emph{arXiv preprint arXiv:2204.06092}.

\bibitem[{Touvron et~al.(2023)Touvron, Martin, Stone, Albert, Almahairi, Babaei, Bashlykov, Batra, Bhargava, Bhosale et~al.}]{touvron2023llama}
Hugo Touvron, Louis Martin, Kevin Stone, Peter Albert, Amjad Almahairi, Yasmine Babaei, Nikolay Bashlykov, Soumya Batra, Prajjwal Bhargava, Shruti Bhosale, et~al. 2023.
\newblock Llama 2: Open foundation and fine-tuned chat models.
\newblock \emph{arXiv preprint arXiv:2307.09288}.

\bibitem[{Wang et~al.(2022)Wang, Wei, Schuurmans, Le, Chi, Narang, Chowdhery, and Zhou}]{wang2022self}
Xuezhi Wang, Jason Wei, Dale Schuurmans, Quoc Le, Ed~Chi, Sharan Narang, Aakanksha Chowdhery, and Denny Zhou. 2022.
\newblock Self-consistency improves chain of thought reasoning in language models.
\newblock \emph{arXiv preprint arXiv:2203.11171}.

\bibitem[{Wei et~al.(2022)Wei, Wang, Schuurmans, Bosma, Xia, Chi, Le, Zhou et~al.}]{wei2022chain}
Jason Wei, Xuezhi Wang, Dale Schuurmans, Maarten Bosma, Fei Xia, Ed~Chi, Quoc~V Le, Denny Zhou, et~al. 2022.
\newblock Chain-of-thought prompting elicits reasoning in large language models.
\newblock \emph{Advances in Neural Information Processing Systems}, 35:24824--24837.

\bibitem[{Xiong et~al.(2022)Xiong, Wan, Hu, Yang, and Li}]{xiong2022self}
Jing Xiong, Zhongwei Wan, Xiping Hu, Min Yang, and Chengming Li. 2022.
\newblock Self-consistent reasoning for solving math word problems.
\newblock \emph{arXiv preprint arXiv:2210.15373}.

\end{thebibliography}
\appendix

\section{Appendix}\label{appendix}
\label{sec:appendix}

\subsection{Runtime Details}\label{ap:implement_details}
We followed \cite{gao2023enabling} to generate 50 responses from ChatGPT and Llama. We used four 48gb A6000 gpus for all experiments. Generating responses using ChatGPT was much faster and only took ~3hrs per dataset. Generating the responses with Llama was much more challenging and took ~24 hrs per dataset.

As \asc\ only contains simple clustering steps, it runs fairly fast with an average of ~3hrs per dataset with ChatGPT. \asc\ with Llama includes the final summarizarion step which took ~15 hrs on average over datasets.

\subsubsection{Tasks}
\asc\ did not use the training set of any of these datasets.\\
\textbf{ASQA} \cite{stelmakh2022asqa}: ASQA is a long-form factoid dataset comprising ambiguous questions. The ambiguous nature of the questions requires answers to comprise diverse facts from multiple documents. We evaluated \asc\ on the eval set with 948 examples.

\noindent\textbf{QAMPARI} \cite{rubin2022qampari}: QAMPARI is a list-style factoid QA dataset constructed from Wikipedia knowledge graphs and tables with the questions paraphrased by humans. We evaluated \asc\ on the eval set with 1000 examples

\noindent\textbf{QUEST} \cite{malaviya2023quest}: QUEST is another list-style dataset constructed using Wiki category lists. We evaluated \asc\ on the test set with 1727 examples 

\noindent\textbf{ELI5} \cite{fan2019eli5}: This is a long-form QA dataset containing how/why/what questions from Reddit. We evaluated \asc\ on the eval split containing 1000 examples.

\begin{figure}[ht]     
     \centering
     \includegraphics[width=\columnwidth]{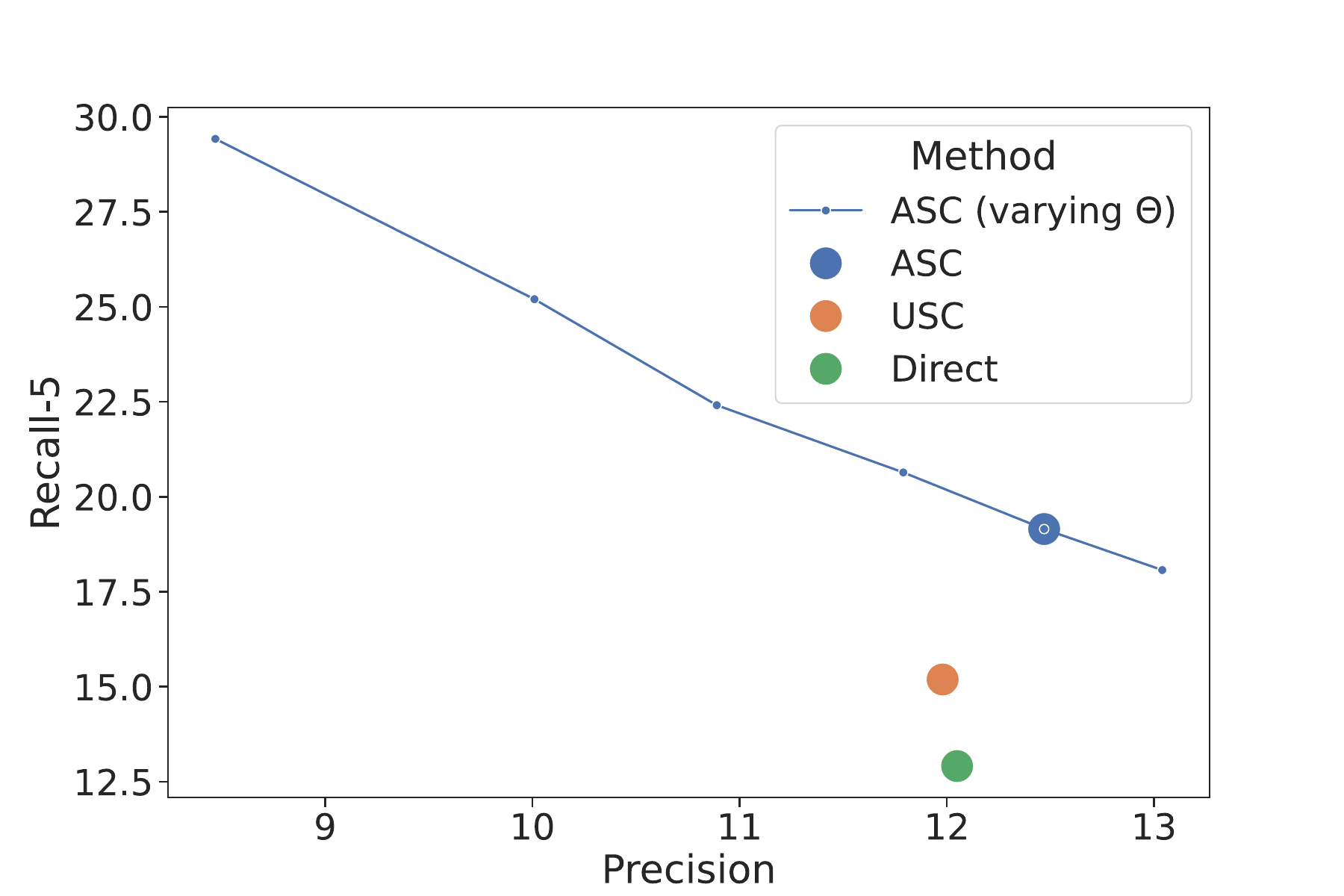}
    \caption{QAMPARI. Increasing $\Theta$ improves precision, reduces recall. Adjusting $\Theta$ produces preferred answer.}
    \label{fig:qampari_vary}
\end{figure}

\begin{figure}[ht]     
     \centering
     \includegraphics[width=\columnwidth]{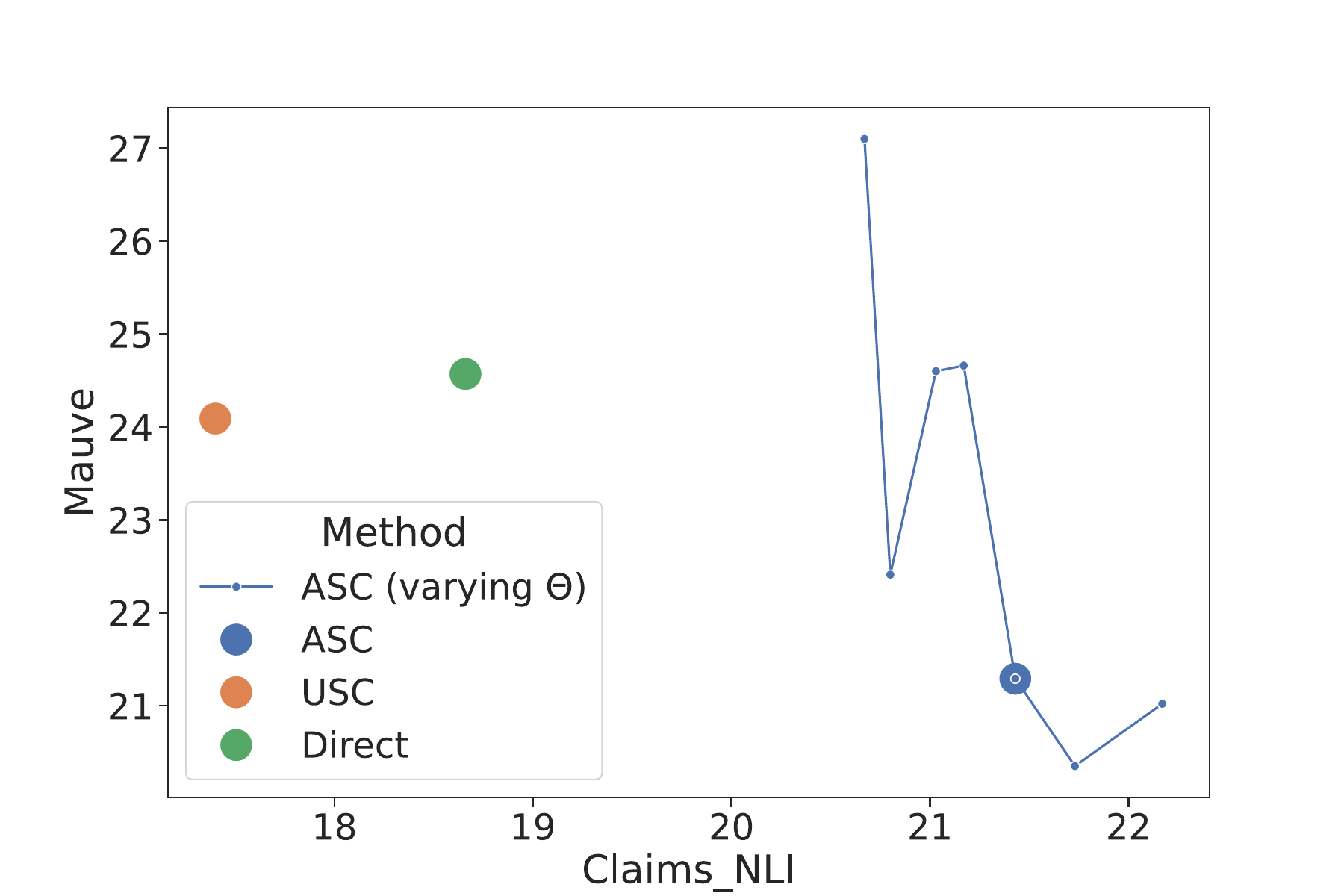}
    \caption{ELI5. Increasing $\Theta$ improves QA-F1 and reduces Mauve. Adjusting $\Theta$ produces preferred answer.}
    \label{fig:eli5_vary}
\end{figure}

\begin{figure}     
     \centering
     \includegraphics[width=\columnwidth]{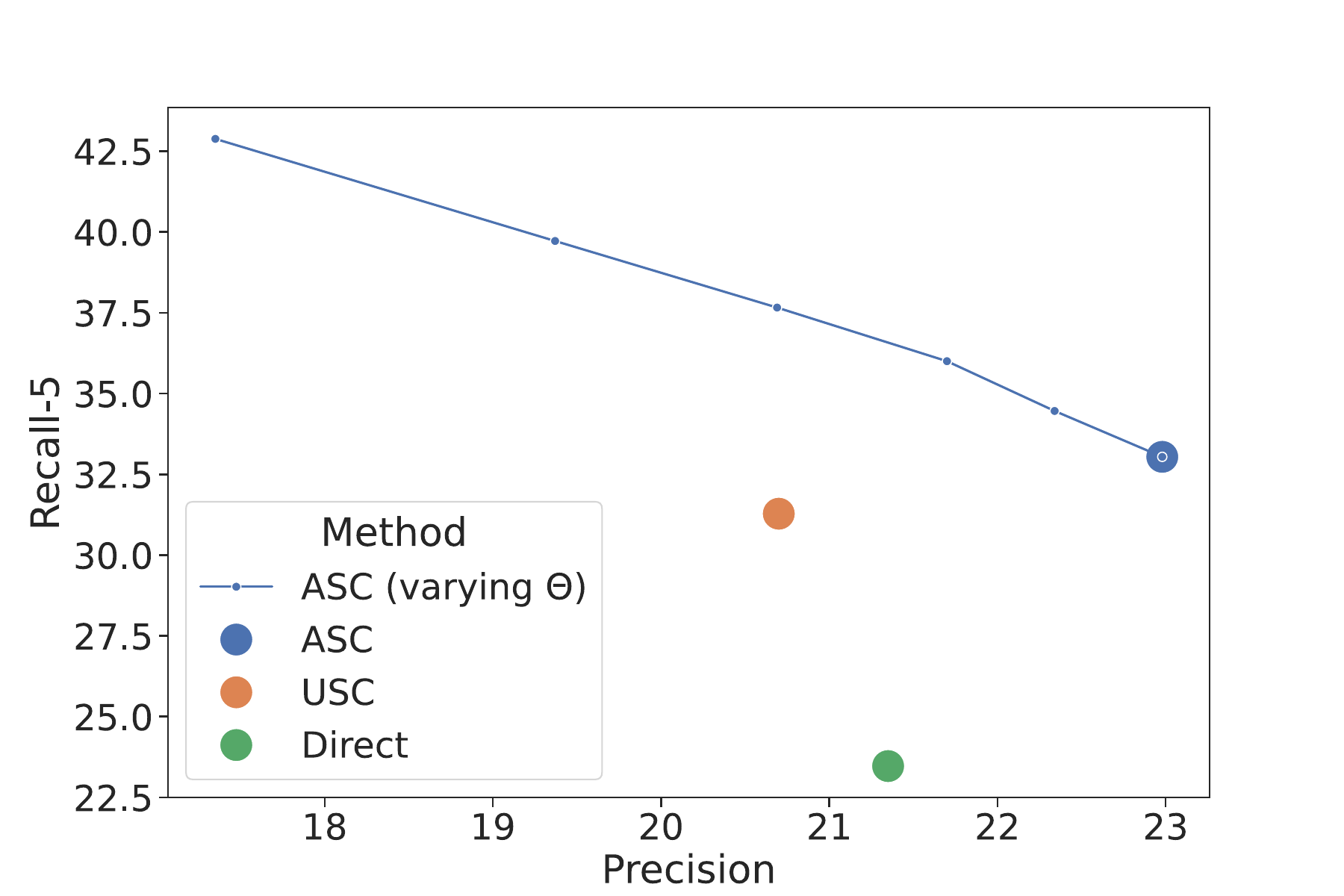}
    \caption{QUEST. Increasing $\Theta$ improves precision, reduces recall. Adjusting $\Theta$ produces preferred answer.}
    \label{fig:quest_vary}
\end{figure}

\subsection{Generating oracle Numbers}\label{ap:orn}
In both ASQA and QAMPARI, we have access to reference short answers. Evaluation metrics - QA\_F1 and precision, recall are all built over these short answers first and then averaged over the entire dataset. For each of these short answers, we find the maximum possible metric value among the 50 generations. This maximum value per short answer is averaged over the entire dataset to get the oracle numbers. For Fig. \ref{fig_pickvsmerge}, we use maximum values at the entire response level rather than at a short answer level.

\subsection{Prompts}\label{ap:prompts}
\subsubsection{Generation Prompts}
We used the exact same generation prompts and few shot exemplars from \cite{gao2023enabling} for ASQA, QAMPARI, ELI5. For QUEST (not analysed by \cite{gao2023enabling}), we used the same prompt as QAMPARI. 
\subsection{Summarization Prompt \texorpdfstring{$P_{combine}$}{P}}Summarization prompts followed the template shown below. An example for asqa summarization with few shot examples is shown later. We used two shot summarization for both ASQA and ELI5.
\begin{tcolorbox}[title=Summarization Template]
\small
\{task instruction\}\\
Question: \{....\}\\
Sentence1: \{...\}\\
Sentence2: \{...\}\\
...\\
Answer: \{....\}
\\ \\
\{task instruction\}\\
Question: \{....\}\\
Sentence1: \{...\}\\
Sentence2: \{...\}\\
...\\
Answer: \{....\} \\ \\
\{task instruction\}\\
Question: \{ ....\}\\
Sentence1: \{...\}\\
Sentence2: \{...\}\\
...\\
Answer:
\end{tcolorbox}

\begin{tcolorbox}[title=Summarization Prompt Example]
\small
Instruction: You are given an ambiguous question and a few sentences which have some parts of its answer and some irrelevant content. Remove irrelevant sentences and combine all relevant ones i
nto a single answer that can address all interpretations of the question. Do not miss any minor details relevant to the question. Also, add any missing details.\\
Question: Where did bruno live in the boy in the striped pajamas?\\
Sentence1: Bruno lived in Germany\\
Sentence2: Bruno moves to Auschwitz when his father got promoted.\\
Sentence3: He is upset about the move.\\
Sentence4: Bruno liked playing with his friends.\\                 
Sentence5: Bruno lives in Berlin.\\               
Sentence6: Bruno discovered a concentration camp near his new home.\\
Sentence7: Bruno was an innocent boy.\\
Answer: Bruno lived in Berlin in Nazi Germany during World War II. His father Ralf gets promoted, and relocates the family to Auschwitz (occupied Poland).\\\\                            
                                                               
Instruction: You are given an ambiguous question and a few sentences which have some parts of its answer and some irrelevant content. Remove irrelevant sentences and combine all relevant ones i
nto a single answer that can address all interpretations of the question. Do not miss any minor details relevant to the question. Also, add any missing details. 

Question: Who played nathan on young and the restless? 
Sentence1: Randy Brooks played nathan on young and restless.\\ 
Sentence2: It was played by Lazarre-White in 1994.\\     
Sentence3: He did an amazing job.\\
Sentence4: Brooks played Nathan.\\
Sentence5: From 1992, Brooks played nathan.\\
Sentence6: He was much younger to his predecessors.\\
Sentence7: He was much younger to his predecessors.\\
Sentence8: Audience liked nathan's portrayal.\\                                                
Answer: The role was played by Nathan Purdee from 1984 to 1992. Randy Brooks took over in 1992 but was replaced in 1994 with a younger version of the character, played by Adam Lazarre-White.\\\\

Instruction: You are given an ambiguous question and a few sentences which have some parts of its answer and some irrelevant content. Remove irrelevant sentences and combine all relevant ones into a single answer that can address all interpretations of the question. Do not miss any minor details relevant to the question. Also, add any missing details.\\
Question: What's the marketing strategy of skipping a number in a numbered line of products?\\
\end{tcolorbox}

\end{document}